\documentclass[10pt,letterpaper,twocolumn]{article}
\usepackage[latin9]{inputenc}
\usepackage{color}
\usepackage{array}
\usepackage{multirow}
\usepackage{amsmath}
\usepackage{amssymb}
\usepackage{graphicx}
\usepackage{subfigure}
\usepackage[unicode=true,
 bookmarks=false,
 breaklinks=true,pdfborder={0 0 1},backref=section,colorlinks=false]
 {hyperref}

\makeatletter

\pdfpageheight\paperheight
\pdfpagewidth\paperwidth

\providecommand{\tabularnewline}{\\}

\usepackage{cvpr}
\usepackage{times}
\usepackage{epsfig}
\usepackage{times}
\usepackage{url}
\usepackage{multirow}

\cvprfinalcopy 
\makeatother

\begin{document}

\title{Sketch-BERT: Learning Sketch Bidirectional Encoder Representation from Transformers by Self-supervised Learning of Sketch Gestalt}

\author{Hangyu Lin$^{*}$, Yanwei Fu
\thanks{indicates equal contributions, $^\dagger$ indicates corresponding author. Y. Fu is with School of Data Science, and MOE Frontiers Center for Brain Science, Shanghai Key Lab of Intelligent Information Processing Fudan University.}\\
School of Data Science, Fudan University\\
{\tt\small {18210980008,yanweifu}@fudan.edu.cn}
\and
Yu-Gang Jiang$^\dagger$, Xiangyang Xue \\
School of Computer Science, Fudan University\\
{\tt\small {ygj, xyxue}@fudan.edu.cn}
}

%

\maketitle

\begin{abstract}
Previous researches of sketches often considered sketches in pixel format and leveraged CNN based models in the sketch understanding. Fundamentally, a sketch is   stored as a sequence of data points, a vector format representation, rather than  the photo-realistic image of pixels.
SketchRNN \cite{ha2018a} studied a generative neural representation for sketches
of vector format by  Long Short Term Memory
networks (LSTM). Unfortunately, the representation learned by SketchRNN is primarily for the generation tasks, rather than the other tasks of recognition and retrieval of sketches.
To this end and inspired by the recent BERT model \cite{devlin2018bert}, we present a model of
learning
Sketch Bidirectional Encoder Representation from Transformer (Sketch-BERT).
We generalize BERT to sketch domain, with the novel proposed components and pre-training algorithms, including the newly designed sketch embedding networks, and  the self-supervised learning of sketch
gestalt. Particularly, towards the pre-training task, we present a novel Sketch Gestalt Model (SGM) to help train the Sketch-BERT. Experimentally, we show that the learned representation of Sketch-BERT can help and improve the performance of the downstream tasks of sketch recognition, sketch retrieval, and sketch gestalt.
\end{abstract}
\begin{figure*}
\centering{}\includegraphics[scale=0.28]{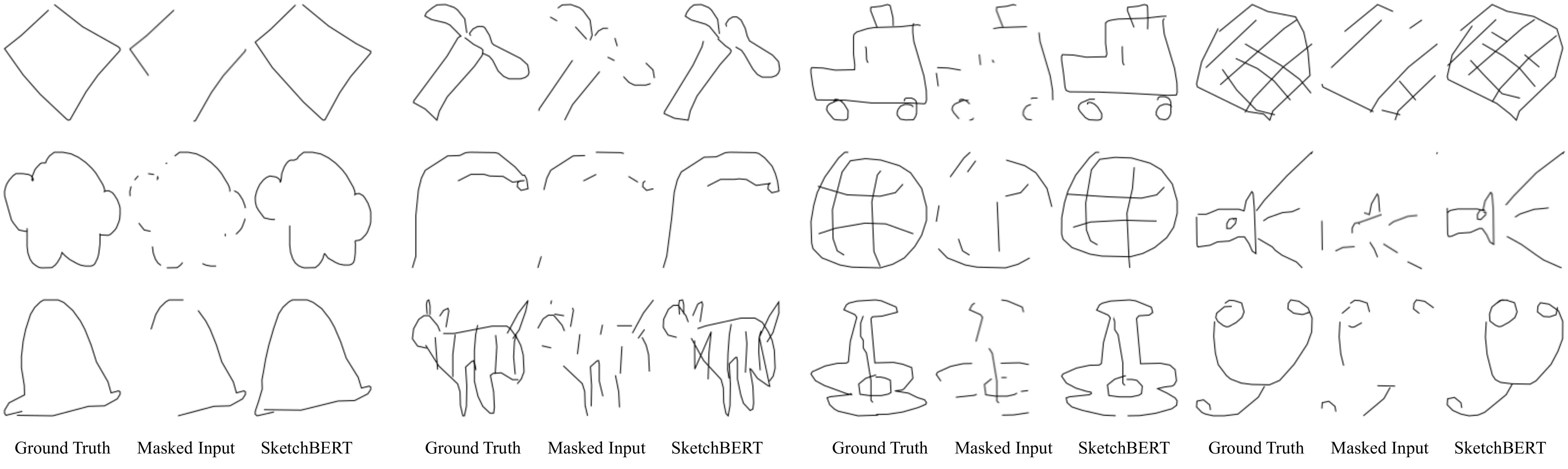} \caption{\label{figs:teaser}Sketch Gestalt which aims at recovering the masked parts of points in sketches and complete the shape of masked sketches.}
\end{figure*}
\section{Introduction}

With the prevailing of touch-screen devices, \emph{e.g.},
iPad, everyone can easily draw simple sketches. It thus supports the
demand of automatically understanding the sketches, which have been
extensively studied in \cite{yu2017sketch,song2017deep,li2018sketch}
as a type of 2D pixel images. Interestingly, the free-hand
sketches reflect our abstraction and iconic representation that
are composed of patterns, structure, form and even simple logic of objects and scenes in the world around us. Thus rather than being
taken as 2D images, sketches should be intrinsically analyzed
from the view of sequential data, which however,
has less been touched in earlier works.
Typically, a sketch consists of several strokes where each stroke can be seen as a sequence of points. We take the same 5-element vector format representation for sketches as in \cite{ha2018a}. Briefly speaking, each point  has 2-dimensional continuous position value and 3-dimensional one hot state value which indicates the state of the point.

According to Gestalt principles of perceptual grouping \cite{gestalt_2004},
humans can easily perceive a sketch as a sequence of data points. To analyze the sequential sketch drawings,
SketchRNN \cite{ha2018a} aimed at learning neural
representation of sketches by combining variational autoencoder
(VAE) with a  Long Short Term Memory networks (LSTM), primary for the sketch generation. In contrast, human vision systems would be capable of both understanding semantics, or abstracting the patterns from sketches.
For instance, we can easily both predict the category label of sketches from ``Ground Truth" column (sketch recognition task), and complete the ``Masked Input" column of sketches (sketch gestalt task), as shown in Fig.~\ref{figs:teaser}.
Comparably, this  demands significant  high quality in learning much more general and comprehensive sketch representation.

Formally, a new sketch Gestalt (sGesta) task is, for the first time, proposed in this paper as in Fig.~\ref{figs:teaser}. The name \emph{sketch Gestalt} comes from the famous Gestalt theory which emphasizes the whole structure of an object rather than some parts. Particularly, the task of sketch gestalt   aims at recovering the masked parts of points in sketches and completes the shape of masked sketches. It  needs to predict both continuous position values and discrete state values which are utilized to define the sketch points. We show that leveraging the sketch gestalt task helps better  understanding the general patterns of sketches.

To this end,  this paper proposes a novel model of  learning  Sketch  Bidirectional  Encoder  Representation from Transformer (Sketch-BERT), which is inspired by the recent BERT model \cite{devlin2018bert} from Natural Language Processing (NLP). Essentially, the transformer structure exerts great potential in modeling the sequential data; and we adopt the weight-sharing multi-layer transformer structure from \cite{lan2019albert}, which share the merits of BERT and yet with much less total parameters. Particularly, a novel embedding method is tailored for sketches, and encodes three level embeddings, \ie, point, positional, and stroke embedding. A refinement embedding network is utilized to project the  embedding features into the input feature space of transformer.

To efficiently train our Sketch-BERT, we introduce a novel task -- self-supervised learning by sketch gestalt, which includes the targets of mask position prediction, and mask state prediction. Correspondingly,   we further present in addressing these tasks,  a novel Sketch Gestalt Model (SGM), which is inspired by  the Mask Language Model in NLP.
The pre-trained Sketch-BERT is capable of efficiently solving the learning tasks of sketches. Particularly, this paper considers the tasks of sketch recognition, sketch retrieval, and sketch gestalt.

 \vspace{0.05in}
\noindent \textbf{Contributions}.
We make several contributions in this paper. (1)
The BERT model is extended to sketches, that is, we for the first time, propose a Sketch-BERT model in efficiently learning neural representation of sketches. Critically, our Sketch-BERT has several novel components, which are significant different from the BERT model, including the novel three-level embedding for sketches, and self-supervised learning by sketch gestalt.
(2) To the best of our knowledge, a novel task -- sketch Gestalt (sGesta) is for the first time studied in this paper. This task is inspired by the Gestalt principles of perceptual grouping. (3) A self-supervised learning process by sketch gestalt, is presented. Empirically, we show that the corresponding SGM for this task can efficiently help pre-train our Sketch-BERT, and thus significantly boost the performance of several downstream sketch tasks.

\section{Related Works}

\noindent  \textbf{Representation of Sketches.} The research on representation of
sketches has been lasted for a long time. As the studies of images
and texts, learning discriminative feature for sketches is also a
hot topic for learning sketch representation. The majority of such
works \cite{hu2011bag,li2013sketch,yu2017sketch,yu2016sketch,lin2019tcnet,li2018sketch}
achieved the goal through the classification or retrieval tasks. Traditional
methods always focused on hand-crafted features,
such as BoW \cite{hu2011bag}, HOG \cite{hu2013performance} and ensemble
structured features \cite{li2013sketch}. Recently, there are works that tried
to learn neural representation of sketches. Due to the  huge visual gap
between sketches and images,
Sketch-A-Net \cite{yu2017sketch} designed a specific Convolutional
Neural Network (CNN) structure for sketches, which achieved the state-of-art
performance at that time, with several following works \cite{yu2016sketch,song2017deep}. On the other hand, TC-Net \cite{lin2019tcnet}
utilized an auxiliary classification task to directly solve the sketch recognition by the backbone, \emph{e.g.},  DenseNet \cite{huang2017densely}. Different from the above methods which directly utilized the pixel level information from sketch images, researchers made use of vector form representation
of sketches in \cite{li2018sketch,zhang2017drawing}.

\vspace{0.05in}
\noindent  \textbf{Generation and  Gestalt of Sketch.} 
Sketch generation, as another significant topic for learning sketches,
also draws more and more attention. 
In \cite{isola2017image,zhu2017unpaired,li2019im2pencil}, they generated
sketches from images via convolutional neural networks and translation
losses. SketchRNN \cite{ha2018a} employed LSTM to solve both conditional
and unconditional generation on vector images of sketches. Reinforcement
learning-based models \cite{zhou2018learning,huang2019learning} also
worked well on learning stroke-wise representation from pixel images
of sketches. Besides the generation task, we propose a new sketch gestalt task in this paper. Despite this task shares the same goal as image inpainting in completing the masked regions/parts, the key differences come from several points, including, (1)
the models for image inpainting \cite{yu2019free,yu2018generative} mostly predict pixels by existing parts in images; in contrast, sketch gestalt  aims at recovering the abstract shapes of some objects.
(2) the texture, color and background information are utilized to help image inpainting models maintain the visual consistency of whole images, while more abstract information, e.g., shape, would be more advisable for sketches in completing the abstraction and iconic sketches.

\vspace{0.05in}
\noindent  \textbf{Transformers and Self-supervised Learning.}
Beside CNN models, it is essential to learn sequence models for learning
how to represent sketches. Recurrent neural networks \cite{hochreiter1997long,chung2014empirical}
are the most successful sequential models during the last decades.
Recently, researchers believe that ``attention is all your need" \cite{vaswani2017attention}; and the  models based on Transformer  are dominating the performance on almost all NLP tasks. Particularly,
 BERT \cite{devlin2018bert}
exploited the mask language model as pre-training task. Further XLNet \cite{yang2019xlnet} generalized
the language modeling strategy in BERT. Such models are all trained in a self-supervised way and then fine-tuned on several downstream tasks.
Inspired by this, we design a novel self-supervised learning method for sketches which can help Sketch-BERT understand the structure of sketches.The task of  self-supervised
learning \cite{self-supervise2019cvpr} is generally defined as learning to predict the withheld parts of data. It thus forces the network to learn what we really care about, such as, image rotation \cite{rotation2018iclr}, image colorization \cite{color2016eccv}, and jigsaw puzzle \cite{ss2018cvpr}. However, most of previous self-supervised learning models are specially designed for images, rather than the sketch. Comparably, the first self-supervised learning by sketch gestalt is proposed and studied in this paper.

. 

\section{Methodology}

This section introduces our Sketch-BERT model and the learning procedure.
Particularly, our model embeds the input sketch as a sequence of
points. A weight-sharing multi-layer transformer is introduced for sketches, and thus it performs as the backbone to our Sketch-BERT. A novel self-supervised learning task -- sketch Gestalt
 task, is proposed to facilitate training Sketch-BERT.

\subsection{Embedding Sketches }

\label{sec:input_embedding} Generally, a sketch is stored as a sequential
set of strokes, which is further represented as a sequence of points.
As the vector data format in \cite{ha2018a}, a sketch
can be represented as a list of points, where each point contains
5 attributes,
\begin{equation}
\left(\Delta x,\Delta y,p_{1},p_{2},p_{3}\right)\label{eq:sketch-data}
\end{equation}
where $\Delta x$ and $\Delta y$ are the values of relative offsets
between current point and previous point; $\left(p_{1},p_{2},p_{3}\right)$
would be utilized as a one-hot vector indicating the state of each
point ($\sum_{i=1}^{3}p_{i}=1$); $p_{2}=1$ indicates the ending
of one stroke; $p_{3}=1$ means the ending of the whole sketch, and
$p_{1}=1$ represents the other sequential points of sketches. We
normalize the position offsets of each point by dividing the maximum
offset values, and make sure $\Delta x,\Delta y\in\left[0,1\right]$.

\vspace{0.1in}

\noindent \textbf{Point Embedding}. Sketches are then embedded
as the sequential representation to learn Sketch-BERT. The point information
$\left(\Delta x,\Delta y,p_{1},p_{2},p_{3}\right)$ is learned as
an embedding
\begin{equation}
\mathrm{E}_{pt}=W_{pt}\left(\Delta x,\Delta y,p_{1},p_{2},p_{3}\right)^{T}\label{eq:point}
\end{equation}
where $W_{pt}\in R^{d_{E}\times5}$ is the embedding matrix, and $d_{E}$
is the dimension of the point embedding.

\vspace{0.1in}
\noindent  \textbf{Positional Embedding}. The position of each sequential point
should be encoded; and thus we introduce the positional embedding
with learnable embedding weight $W_{ps}$,
\begin{equation}
\mathrm{E}_{ps}=W_{ps}\boldsymbol{1}_{ps}\in R^{d_{E}}\label{eq:position}
\end{equation}
where $\boldsymbol{1}_{ps}$ is one-hot positional vector. In particular,
we set the max length of each sketch sequence up to 250, while remove
the points of the sequence beyond 250, by default.

 \vspace{0.1in}
\noindent  \textbf{Stroke Embedding}. We also learn to embed the sequences of
strokes. Inspired by the segment embedding in language model \cite{devlin2018bert},
the strokes of sketch are also embedded as
\begin{equation}
\mathrm{E}_{str}=W_{str}\boldsymbol{1}_{str}\in R^{d_{E}}\label{eq:stroke}
\end{equation}
with the length of stroke sequence up to 50; where $\boldsymbol{1}_{str}$
is corresponding one-shot stroke vector. Thus, we have the following
final sketch embedding as,
\begin{eqnarray}
\mathrm{E}=\mathrm{E}_{pt}+\mathrm{E}_{ps}+\mathrm{E}_{str}\label{eq:total-embed}
\end{eqnarray}

\noindent \textbf{Refine Embedding Network.}
We further employ a refine embedding network to improve the embedding
dimension from $d_{E}$ to $d_{H}$, used in the transformer. Specifically,
the refine embedding network consists of several fully-connected layers
with the input and output dimensions $d_{E}$ and $d_{H}$, respectively.
In our Sketch-Bert, we have $d_{E}=128,d_{H}=768$, and the structure
of refinement network is $128-256-512-768$, where the neurons of
two hidden layers are 256 and 512, respectively.

\subsection{Weight-sharing Multi-layer Transformer \label{subsec:Weight-sharing-Multi-layer-Trans}}


\begin{figure*}
\centering{}
\subfigure[Sketch-BERT for Sketch Gestalt]
{\includegraphics[scale=0.28]{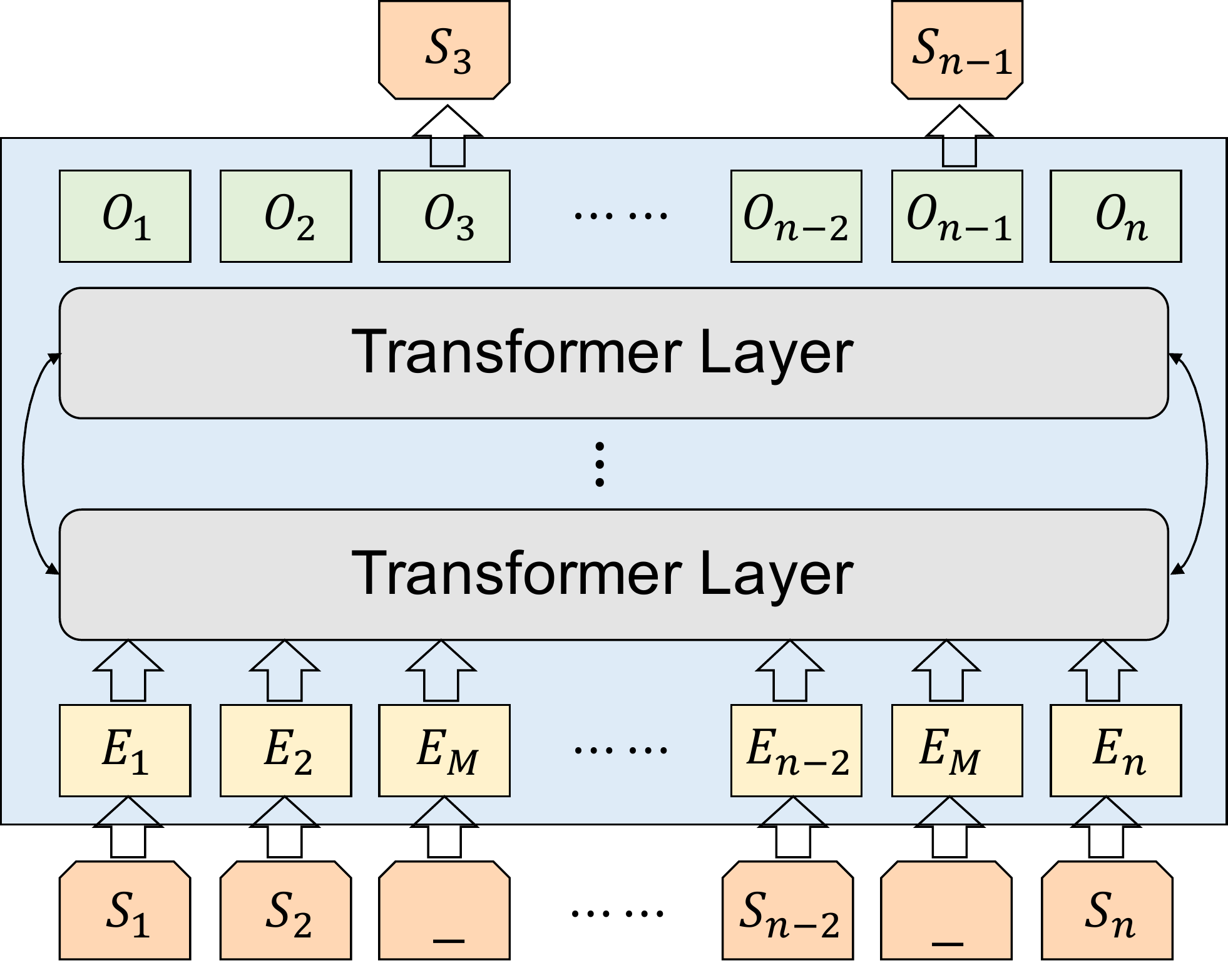}
\label{figs:struct_a}}
\hspace{.9in}
\subfigure[Sketch-BERT for Sketch Recognition/Retrieval]
{\includegraphics[scale=0.28]{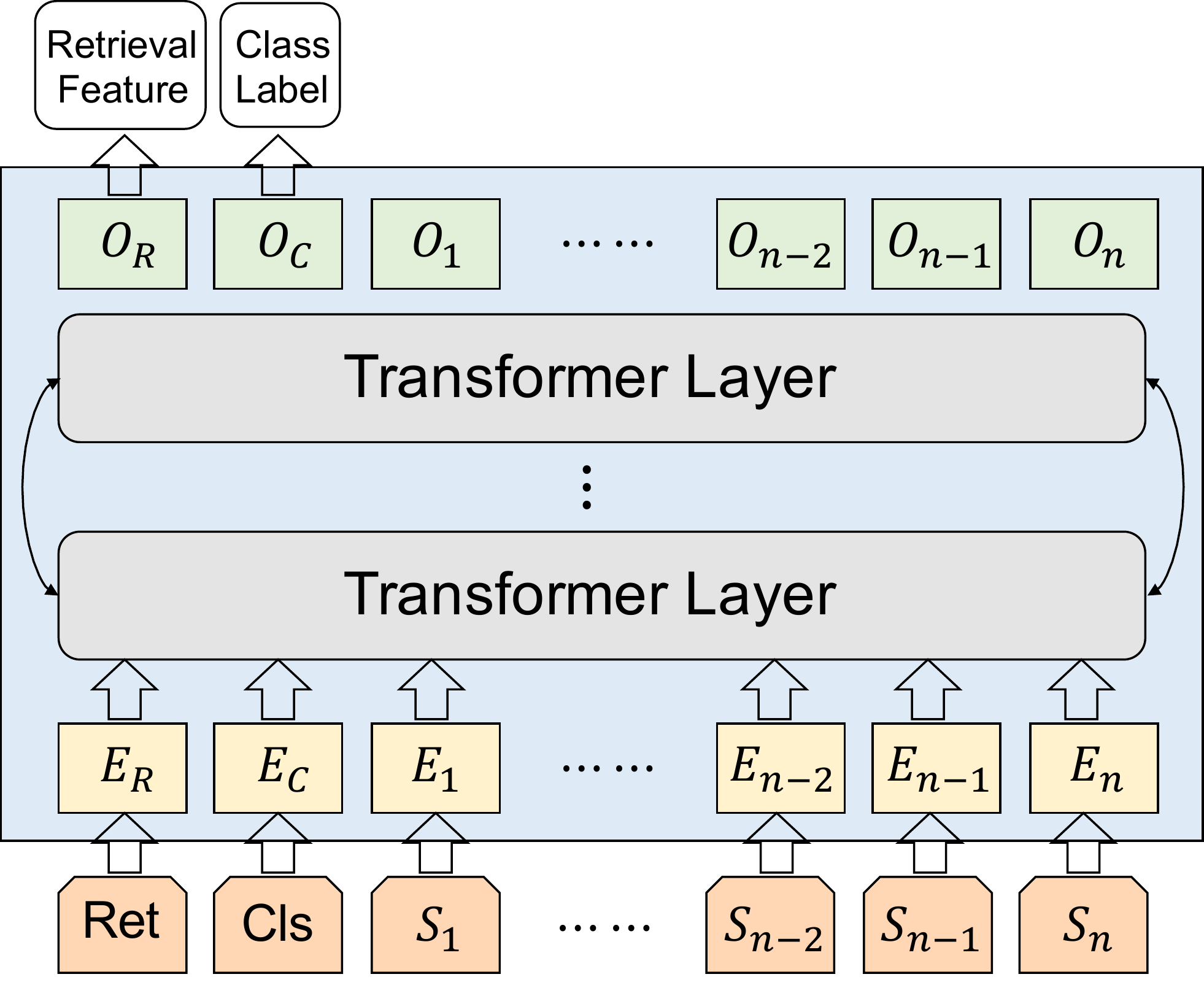}
\label{figs:struct_b}}
\caption{\label{figs:structure}Overview structure of Sketch-BERT for Sketch
Gestalt Model and downstream tasks.}
\end{figure*}

We adopt the weight-sharing multi-layer bidirectional
transformer as the backbone,  inspired by the ALBERT \cite{lan2019albert} and BERT
\cite{devlin2018bert}. Particularly, the weights are shared in the
layers of the encoder. This makes a faster convergence of Sketch-BERT.
Formally, we denote the sketch embedding as
\[
\mathrm{E}=\left(\mathrm{E}_{1},\mathrm{E}_{2},\cdots,\mathrm{E}_{n}\right)\in R^{n\times d_{H}}
\]
where $n$ is the true length of each sketch embedding. Hidden features
will be updated by self-attention module in each weight-sharing transformer
layer. The final output features from the Sketch-BERT encoder will
be used for different downstream tasks.

\subsection{Self-Supervised Learning by Sketch Gestalt}

Since the pre-training tasks over unlabeled text data in NLP have shown great potential in improving
the performance of BERT, it is essential to introduce a self-supervised
learning task to facilitate pre-training our Sketch-BERT.

To this end, we define a novel self-supervised learning process by
sketch Gestalt (sGesta), which aims at recovering the masked points
in sketches as shown in Fig.~\ref{figs:struct_a}. Given a masked sketch in vector format $s_{mask}=s_{gt}\cdot m$ where $m$ is the mask with the same shape of $s_{gt}$, sketch Gestalt targets at predicting $s_{comp}$ which has the same shape and semantic information as $s_{gt}$ from the $s_{mask}$. Specifically, the position mask at first two dimensions and state mask at other dimensions can be predicted, individually. To solve the  self-supervised learning task, we present the \emph{Sketch Gestalt
Model} (SGM). As in Eq
(\ref{eq:sketch-data}), each point is represented by the key information
of both positional offset $(\Delta x,\Delta y)$ and state $\left(p_{1},p_{2},p_{3}\right)$,
which will be masked and predicted by our SGM, individually. We propose different mask strategies for positional offset and state information to help train our
Sketch-BERT. By default, we mask $15\%$ of all positions and states
respectively for each sketch sequence.

 \vspace{0.05in}
 \noindent  \textbf{Mask Position Prediction}. We divide the offset values for points into two
classes: 1) the offset for a point in a stroke; 2) the offset for
a point as the start of a stroke. In sketches, distributions of these two type offset values are quite different, and there are also total distinctive value ranges of two types of offset values.
 Thus we generate the masks by sampling points in these two classes, proportional to the total point
number of each point type class, by setting $(\Delta x,\Delta y)$
of the masked point to $0$.

 \vspace{0.05in}
\noindent  \textbf{Mask State Prediction.} Quite similarly, there are imbalance
distributions of $p_{1},p_{2},p_{3}$ for sketch points. In particular,
there are always much more points with $p_{1}$ than those with $p_{2}$
or $p_{3}$. Thus, we mask the state of each point, in term of the
percentage of points with the state $p_{1},p_{2},p_{3}$. If the state
of one point is masked, it has $p_{1}=p_{2}=p_{3}=0$.

 \vspace{0.05in}
\noindent  \textbf{Embedding Reconstruction Network.} Our SGM introduces an embedding
reconstruction network, which plays the corresponding decoder of the
refine embedding network. In particular, given as the input the $d_{H}$
dimensional embedding features, the reconstruction network predicts
the states and positions of each mask. Practically, we reverse the
structure of refine embedding network, and utilize the structure as
$768-512-256-128-5$, with the  neurons of 512, 256, and 128 of hidden layers, individually. We adopt $L_{1}$ loss for mask position prediction,
to predict the continuous position offset values; and, we
use the standard cross entropy loss for different state categories
in mask state prediction.

\subsection{Learning Tasks by Sketch-BERT \label{sec:tasks}}

\noindent We further elaborate how Sketch-BERT model could be utilized
for different downstream tasks after the pre-training procedure by
the self-supervised learning. For each task, we give the formal definition
 and describe how the pre-trained Sketch-BERT model can
be utilized here. Especially, we are interested in following tasks.

 \vspace{0.05in}
\noindent  \textbf{Sketch Recognition}. This task takes a sketch $s$ as input
and predicts its category label $c$. To fine-tune
the Sketch-BERT for recognition task, we add a \textsc{{[}CLS{]}}
label, \emph{i.e.}, a special token to the beginning of the sequential data of each
sketch, as shown in Fig.~\ref{figs:struct_b}. For recognition
tasks, our Sketch-BERT serves as a generic feature extractor of each
sketch. A standard softmax classification layer as well as cross entropy loss, is applied to the
outputs of Sketch-BERT ($O_{C}$). The training sketches of recognition
tasks have been utilized to fine-tune the Sketch-BERT, and train the
classification layer, as the standard practice in BERT \cite{devlin2018bert}.

 \vspace{0.05in}
\noindent  \textbf{Sketch Retrieval.} Given a query sketch $s_{q}$, sketch
retrieval task targets at finding sketches $s_{1},\dots,s_{n}$ with
the same category as the query $s_{q}$. We add the \textsc{{[}RET{]}}
label token to the beginning of sequential data of each sketch, and
use the Sketch-BERT to extract the features ($O_{R}$) of each sketch,
as in Fig.~\ref{figs:struct_b}. To conduct the retrieval task,
the output features are projected into a fully connected layer of
256 neurons, which is optimized by a triplet loss as in \cite{lin2019tcnet}
by minimizing the distance of sketches in the same class, and maximizing
the distance of sketches in different classes. In addition, we also apply the
cross entropy loss of learning to predict the category of each sketch.
The training data of
retrieval task is utilized to train the newly added fully connected
layer, and fine-tune the Sketch-BERT.

\vspace{0.1in}
\noindent  \textbf{Sketch Gestalt.} Inspired by the Gestalt principles of perceptual
grouping, this task is introduced to recover a realistic sketch images
$s_{comp}$ given an incomplete $s_{mask}$ as shown in Fig.~\ref{figs:struct_a}. We directly utilize the
SGM learned in self-supervised learning step for this task.

\noindent 

\section{Experiments and Discussion}

\subsection{Datasets and Settings\label{sec:data_set}}

\noindent \textbf{Datasets.} Our model is  evaluated on two  large-scale sketch datasets -- QuickDraw dataset \cite{ha2018a},
and TU-Berlin dataset \cite{eitz2012hdhso} 
(1) QuickDraw dataset is collected from Google application \textit{Quick,
Draw!}, an online game to draw a sketch
less than 20 seconds. There are about 50 million sketch drawings across
total 345 classes of common objects. Here we follow
the pre-process method and training split from \cite{ha2018a}, where
each class has 70K training samples, 2.5K validation and 2.5K test
samples in QuickDraw dataset. We also simplify the sketches by applying
the Ramer-Douglas-Peucker (RDP) algorithm, leading to a maximum sequence
length of 321. (2) TU-Berlin contains less quantity but better quality
sketch samples than QuickDraw. There are 250 object categories in
TU-Berlin with 80 sketches in each category.

\noindent \textbf{Implementation Details.} In our work, the Sketch-BERT
model has $L=8$ weight-sharing Transformer layers with the hidden
size of $H=768$ and the number of self-attention heads of $12$.
The same with BERT, the feed-forward size will be set to $4H$ in
the weight-sharing transformer layer. The embedding size is set to
$128$ and the refine embedding network is a fully-connected network
of neurons $128-256-512-768$. Correspondingly, the reconstruction
network is composed of four fully-connected layers of neurons $768-512-256-128-5$.
The max lengths of input sketches are set as 250, and 500 for QuickDraw,
and TU-Berlin, respectively. We implement our Sketch-BERT model with
PyTorch. To optimize the whole model, we adopt Adam optimizer with
a learning rate of $0.0001$. In self-supervised learning, we leverage
the whole training data from QuickDraw to train the sketch gestalt
model.
\begin{table}
\begin{centering}
\begin{tabular}{c|c|c|c|c}
\hline
\multirow{2}{*}{{\small{}Methods} } & \multicolumn{2}{c|}{{\small{}{}{}QuickDraw (\%)}} & \multicolumn{2}{c}{{\small{}{}{}TU-Berlin (\%)}}\tabularnewline
\cline{2-5}
 & {\small{}{}{}T-1}  & {\small{}{}{}T-5}  & {\small{}{}{}T-1}  & {\small{}{}{}T-5}\tabularnewline
\hline
\hline
{\small{}{}{}HOG-SVM \cite{eitz2012hdhso}}  & {\small{}{}{}56.13 }  & {\small{}{}{}78.34 }  & {\small{}{}{}56.0 }  & {\small{}{}{}--} \tabularnewline
\hline
{\small{}{}{}Ensemble \cite{li2013sketch}}  & {\small{}{}{}66.98}  & {\small{}{}{}89.32 }  & {\small{}{}{}61.5 }  & {\small{}{}{}-- } \tabularnewline
\hline
{\small{}{}{}Bi-LSTM \cite{hochreiter1997long}}  & {\small{}{}{}86.14 }  & {\small{}{}{}97.03 }  & {\small{}{}{}62.35}  & {\small{}{}{}85.25 } \tabularnewline
\hline
{\small{}{}{}Sketch-a-Net$^{*}$ \cite{yu2016sketch}}  & {\small{}{}{}-- }  & {\small{}{}{}-- }  & {\textbf{\small{}{}{}77.95 }}  & {\small{}{}{}-- } \tabularnewline
\hline
{\small{}{}{}Sketch-a-Net \cite{yu2016sketch}}  & {\small{}{}{}75.33 }  & {\small{}{}{}90.21 }  & {\small{}{}{}47.70 }  & {\small{}{}{}67.00 } \tabularnewline
\hline
{\small{}{}{}DSSA \cite{song2017deep}}  & {\small{}{}{}79.47 }  & {\small{}{}{}92.41 }  & {\small{}{}{}49.95 }  & {\small{}{}{}68.00 } \tabularnewline
\hline
{\small{}{}{}ResNet18 \cite{he2016deep}}  & {\small{}{}{}83.97 }  & {\small{}{}{}95.98 }  & {\small{}{}{}65.15 }  & {\small{}{}{}83.30} \tabularnewline
\hline
{\small{}{}{}ResNet50 \cite{he2016deep}}  & {\small{}{}{}86.03 }  & {\small{}{}{}97.06 }  & {\small{}{}{}69.35 }  & {\small{}{}{}90.75 }\tabularnewline
\hline
{\small{}{}{}TCNet \cite{lin2019tcnet}}  & {\small{}{}{}86.79 }  & {\small{}{}{}97.08 }  & {\small{}{}{}73.95 }  & {\small{}{}{}91.30 }\tabularnewline
\hline
\hline
{\small{}{}Sketch-BERT} {\small{}{}(w./o.)}  & \multicolumn{1}{c|}{{\small{}{}{}{}{}83.10 }} & \multicolumn{1}{c|}{{\small{}{}{}{}{}95.84 }} & \multicolumn{1}{c|}{{\small{}{}{}{}{}54.20 }} & \multicolumn{1}{c}{{\small{}{}{}{}{}66.05 }}\tabularnewline
\hline
{\small{}{}Sketch-BERT} {\small{}{}(w.)}  & \multicolumn{1}{c|}{\textbf{\small{}{}{}{}{}88.30}} & \multicolumn{1}{c|}{\textbf{\small{}{}{}{}{}97.82}{\small{}{}{}{}{} }} & \multicolumn{1}{c|}{{\small{}{}{}{}{}76.30}{\small{}{}{}{}{} }} & \multicolumn{1}{c}{\textbf{\small{}{}{}{}{}91.40}{\small{}{}{}{}{} }}\tabularnewline
\hline
\end{tabular}
\par\end{centering}
\caption{\label{tab:res_cls}The Top-1 (T-1) and Top-5 (T-5) accuracy of our
model and other baselines on classification task; w./o., and w. indicate
the results without, and with the self-supervised learning by sketch
Gestalt, individually. $^{*}$ means the results in original paper \cite{yu2016sketch}.}
\end{table}

\noindent \textbf{Competitors.} We compare several baselines here.
(1) HOG-SVM \cite{eitz2010sketch}: It is a traditional method utilized HOG feature and SVM to predict the classification result.  (2) Ensemble \cite{li2013sketch}: This model leverages several types of features for sketches, we evaluate it on classification task. (3) Bi-LSTM \cite{hochreiter1997long} : We employ a three-layer bidirectional LSTM model to test the recognition
and retrieval tasks on sequential data of sketches. The dimension
of the hidden states is set to 512 here. (4) Sketch-a-Net: \cite{yu2017sketch}:
The Sketch-a-Net is a specifically designed convolutional neural network
for sketches. (5) DSSA\cite{song2017deep} add an attention module and a high-order energy triplet loss function to original Sketch-A-Net model. (6) ResNet: We also evaluate residual network, one of
the most popular convolutional neural network in computer vision field
designed for image recognition task. (7) TC-Net \cite{lin2019tcnet}:
It is a network based on DenseNet \cite{huang2017densely} for sketch
based image retrieval task, we leverage the pre-trained model for classification and retrieval tasks. (8) SketchRNN \cite{ha2018a}: SketchRNN employed a variational autoencoder with LSTM network as encoder and decoder backbones to solve the sketch generation task, in our experiments, we use this approach to test the sketch gestalt task.
The training and validation set of datasets are employed to train our models and competitors, which are further validated in the test set.
For fair comparison of structure, we retrain all models on QuickDraw and TU-Berlin datasets for different tasks.

\begin{table*}
\centering{}%
\begin{tabular}{c|c|c|c|c|c|c}
\hline
\multirow{2}{*}{{\small{}Models} } & \multicolumn{3}{c|}{{\small{}{}{}QuickDraw}} & \multicolumn{3}{c}{{\small{}{}{}TU-Berlin}}\tabularnewline
\cline{2-7}
 & {\small{}{}{}Top-1 (\%)}  & {\small{}{}{}Top-5 (\%)}  & {\small{}{}{}mAP (\%) }  & {\small{}{}{}Top-1 (\%)}  & {\small{}{}{}Top-5 (\%)}  & {\small{}{}{}mAP(\%) } \tabularnewline
\hline
\hline
{\small{}{}{}Bi-LSTM \cite{hochreiter1997long}}  & {\small{}{}{}70.91 }  & {\small{}{}{}89.52 }  & {\small{}{}{}60.11 }  & {\small{}{}{}31.40 }  & {\small{}{}{}59.60 }  & {\small{}{}{}23.71 }\tabularnewline
\hline
{\small{}{}{}Sketch-a-Net \cite{yu2016sketch} }  & {\small{}{}{}74.88 }  & {\small{}{}{}90.10 }  & {\small{}{}{}65.13 }  & {\small{}{}{}37.25 }  & {\small{}{}{}63.50 }  & {\small{}{}{}26.18} \tabularnewline
\hline
{\small{}{}{}DSSA \cite{song2017deep} }  & {\small{}{}{}78.16 }  & {\small{}{}{}91.04 }  & {\small{}{}{}68.10 }  & {\small{}{}{}38.45 }  & {\small{}{}{}66.10 }  & {\small{}{}{}28.77} \tabularnewline
\hline
{\small{}{}{}ResNet18 \cite{he2016deep} }  & {\small{}{}{}80.34 }  & {\small{}{}{}91.71 }  & {\small{}{}{}70.98 }  & {\small{} 41.45}  & {\small{}{}{}67.10}  & {\small{}{}{}29.33 } \tabularnewline
\hline
{\small{}{}{}ResNet50 \cite{he2016deep}}  & {\small{}{}{}82.41 }  & {\small{}{}{}92.52 }  & {\small{}{}{}74.84 }  & {\small{}{}{}51.80 }  & {\small{}{}{}74.45}  & {\small{}{}{}36.94} \tabularnewline
\hline
{\small{}{}{}TCNet \cite{lin2019tcnet}}  & {\small{}{}{}83.59 }  & {\small{}{}{}92.57 }  & {\small{}{}{}76.38 }  & {\small{}{}{}55.30}  & {\small{}{}{}79.45 }  & {\small{}{}{}38.78 }\tabularnewline
\hline
\hline
{\small{}Sketch-BERT} {\small{}(w./o.)} & \multicolumn{1}{c|}{{\small{}{}{}63.13 }} & \multicolumn{1}{c|}{{\small{}{}{}84.70 }} & \multicolumn{1}{c|}{{\small{}{}{}55.10 }} & \multicolumn{1}{c|}{{\small{}{}{}32.50 }} & \multicolumn{1}{c|}{{\small{}{}{}57.90 }} & {\small{}{}{} 24.14}\tabularnewline
\hline
{\small{}Sketch-BERT } {\small{}(w.)} & \multicolumn{1}{c|}{\textbf{\small{}{}{}85.47}{\small{}{}{} }} & \multicolumn{1}{c|}{\textbf{\small{}{}{}93.49}{\small{}{}{} }} & \multicolumn{1}{c|}{\textbf{\small{}{}{}78.87}{\small{}{}{} }} & \multicolumn{1}{c|}{\textbf{\small{}{}{}57.25}{\small{}{}{} }} & \multicolumn{1}{c|}{\textbf{\small{}{}{}81.50}{\small{}{}{} }} & \multicolumn{1}{c}{\textbf{\small{}{}{}41.54}{\small{}{}{} }}\tabularnewline
\hline
\end{tabular}\caption{\label{tab:res_ret}The Top-1, Top-5 accuracy and mean Average Precision(mAP) of our model and other baselines on
sketch retrieval task. w./o., and w. indicate the results without,
and with the self-supervised learning by sketch gestalt. }
\end{table*}

\subsection{Results on Sketch Recognition Task}

Recognition or classification is a typical task for understanding
or modeling data in term of semantic information, so we first compare
the classification results of our model with other baselines. We use
100 categories with 5K train samples, 2.5K validation samples and
2.5K test samples for QuickDraw dataset; whole categories of TU-Berlin
dataset with training split of $80\%/10\%/10\%$ for train/validation/test
samples, respectively.

From the results in Tab. \ref{tab:res_cls}, it is obvious that the
Sketch-BERT outperforms other baselines including both pixel images
based models like Sketch-a-Net, ResNet18/50 or TC-Net; and vector
images based model like Bi-LSTM by a considerable margin: about $2\%$
on QuickDraw. This indicates the effectiveness of our Sketch-BERT model, and self-supervised pipeline by sGesta. Particularly, we give the ablation study of our Sketch-BERT without using self-supervised training (i.e., Sketch-BERT (w./o.) in Tab. \ref{tab:res_cls}). It gives us the results of $5\%$ dropping of top-1 accuracy on QuickDraw dataset.  In fact, this can reveal the power of our SGM proposed in this paper. Furthermore, the Sketch-BERT (w.) gets converged much faster than that of  Sketch-BERT (w./o.) if they are fine-tuned on the same training data. For example,  the convergence
epoch reduces from 50 epochs of  Sketch-BERT (w./o.), to only 5 epochs of Sketch-BERT (w.), for the recognition task trained on TU-Berlin dataset.


\subsection{Results on Sketch Retrieval Task}

We are particularly  interested in  the category-level sketch retrieval and test sketch retrieval task over the same dataset  as the recognition task.
To evaluate the performance of different models, we report
both Top-1/5 accuracy and mean Average Precision (mAP). To make a fair comparison to the other baselines
We employ the typical triplet loss and cross entropy loss, as our Sec. \ref{sec:tasks}.  Each  model only serves as the backbone to extract the sketch features from the a tuple of   anchor sketch, positive sketch, negative
sketch. The ranked retrieval results are compared.



The results are summarized in Tab. \ref{tab:res_ret}.
Our
Sketch-BERT model with self-supervised learning tasks has a much higher performance
than the other baselines. It gives us  about $2\%$ improvement over the best second method ---
TCNet, which is the state-of-the-art CNN based model for sketch recognition.
We notice that the vector based model -- Bi-LSTM only achieves
 $70\%$ top-1 accuracy, while the others CNN based models get the performance over $75\%$ accuracy. On the other hand, interestingly our Sketch-BERT without self-supervised training by sGesta, achieves much worse results than the other baselines on this retrieval task.  This further suggests that our SGM model proposed in self-supervised learning step, can efficiently improve the generalization ability of our Sketch-BERT. To sum up, the results from both sketch classification and sketch
retrieval tasks show the superiority of our Sketch-BERT model on the sketch representation learning.

\subsection{Results on Sketch Gestalt Task}

\begin{figure*}
\centering{ \includegraphics[scale=0.27]{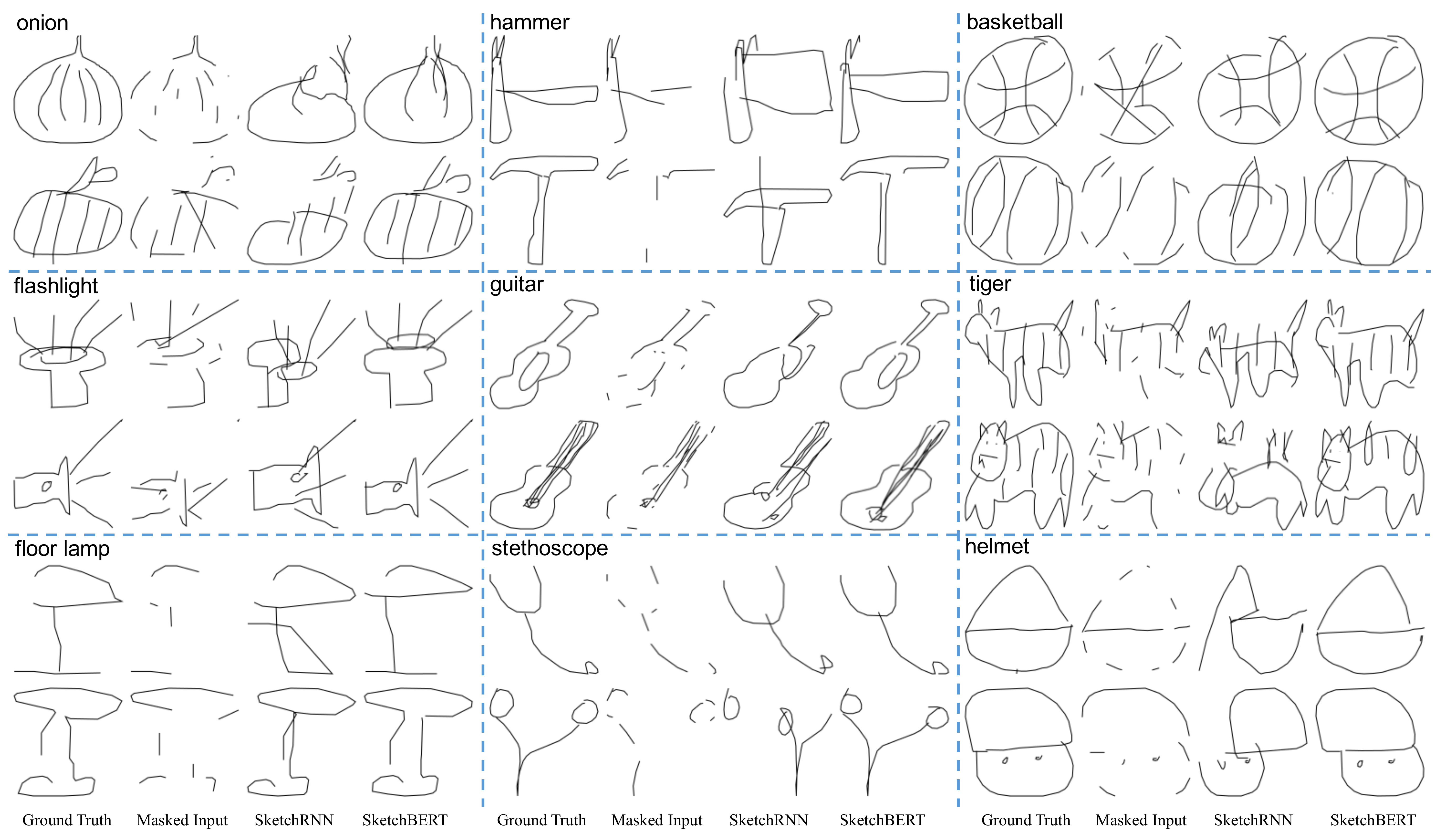} \caption{\label{figs:result_comp} Completion results on sketch gestalt of our Sketch-BERT and SketchRNN on QuickDraw dataset from 9 classes, \emph{onion,flashlight,  floor lamp, hammer, guitar, stethoscope,basketball, tiger,helmet}.}
}
\end{figure*}

Rather than discriminative neural representation, Sketch-BERT model also has a good capacity for generative representation like sketch gestalt task, where some part of sketches have been masked, and predicted by the models. In this section, our model is compared against  SketchRNN \cite{ha2018a}, which, to the best of our knowledge, is the only generative model that is able to predict the masked sketch sequences.  This task is conducted on QuickDraw dataset: both models are learned on training data, and predicted on the test data.

We illustrate some completed results from several classes in QuickDraw dataset in Fig. \ref{figs:result_comp}.
The four columns in the
figure represent (1) ground truth sketch, (2) incomplete or masked input
with a random $30\%$ mask on position and state together, (3) completed
results from the SketchRNN, (4) completed results from our Sketch-BERT
model.

We can show that  our Sketch-BERT model has a much better
ability in understanding and filling the masked sketches  in a more
correct way than that of SketchRNN. Particularly, we further analyze and compare these results.  As for the simple sketches, SketchRNN   has a
reasonable  ability in completing the missing parts of each sketch. For example, we can observe the general good examples  from the first column of  SketchRNN
in  Fig. \ref{figs:result_comp}.  However, SketchRNN is quite limited to
fill the complicated sketches, such as the \emph{flashlight,tiger},
SketchRNN may be failed to complete them. In contrast,  our Sketch-BERT can still correctly
capture both the shape and  details of such sketches as the results in the second and third columns of  Fig. \ref{figs:result_comp}. We also show more examples of different classes
on sketch gestalt task in supplementary material. Besides the qualitative results, we also provide a user study as the quantitative comparison in the supplementary material.

\subsection{Pre-training Task Analysis}

\begin{table}
\centering{}%
\begin{tabular}{c|c|c|c|c}
\hline
\multirow{2}{*}{{\small{}Models} } & \multicolumn{2}{c|}{{\small{}{}{}Classification}} & \multicolumn{2}{c}{{\small{}{}{}Retrieval}}\tabularnewline
\cline{2-5}
 & {\small{}{}{}Top-1 (\%)}  & {\small{}{}{}Top-5 (\%) }  & {\small{}{}{}Top-1 (\%)}  & {\small{}{}{}Top-5(\%) } \tabularnewline
\hline
\hline
{\small{}{}{}Single }  & {\small{}{}{}86.51 }  & {\small{}{}{}96.72 }  & {\small{}{}{}81.73 }  & {\small{}{}{}92.13 } \tabularnewline
\hline
{\small{}{}{}Position}  & {\small{}{}{}87.37 }  & {\small{}{}{}97.01 }  & {\small{}{}{}82.22 }  & {\small{}{}{}91.98 } \tabularnewline
\hline
{\small{}{}{}State }  & {\small{}{}{}86.83 }  & {\small{}{}{}96.88 }  & {\small{}{}{}81.87 }  & {\small{}{}{}92.15} \tabularnewline
\hline
{\small{}{}{}Full }  & \textbf{\small{}{}{}88.30 }{\small{}{} }  & \textbf{\small{}{}{}97.82}{\small{}{} }  & \textbf{\small{}{}{}85.47 }{\small{}{} }  & \textbf{\small{}{}{}93.49}{\small{}{} }\tabularnewline
\hline
\end{tabular}\caption{\label{tab:res_ablation_pre_train} The performance of classification
and retrieval tasks on QuickDraw dataset after different types of pre-training tasks.}
\end{table}

In this section, we give further ablation study and  analyze how the self-supervised learning  and models
can affect the performance on sketch representation learning.

\noindent \textbf{Different Pre-training Tasks.} First, we study the
different pre-training tasks in our model: (1) Single, means the traditional
random mask strategy used in BERT; (2)Position, means that only masks
the position information according to the mask strategy in our sketch
gestalt model; (3)State,  masks the state information, (4) Full, is the full
newly proposed mask strategy in sketch gestalt model. We show the
performance of standard Sketch-BERT on the classification and retrieval
tasks after these pre-training tasks in Tab. \ref{tab:res_ablation_pre_train}.

It is clear that our sketch gestalt model plays an important role to improve the performance of Sketch-BERT, and we notice  there is  a consistent improvement
over the other mask models: Single ($>1.7\%$), Position ($>1\%$), State ($>1.4\%$).
This  reveals the significance of a proper mask model for learning
the good neural representation of sketches. Furthermore, we can find
the position information plays a more important role to
sketch representation learning than the state information, as in Tab. \ref{tab:res_ablation_pre_train}.

\begin{figure}
\centering{}\includegraphics[scale=0.4]{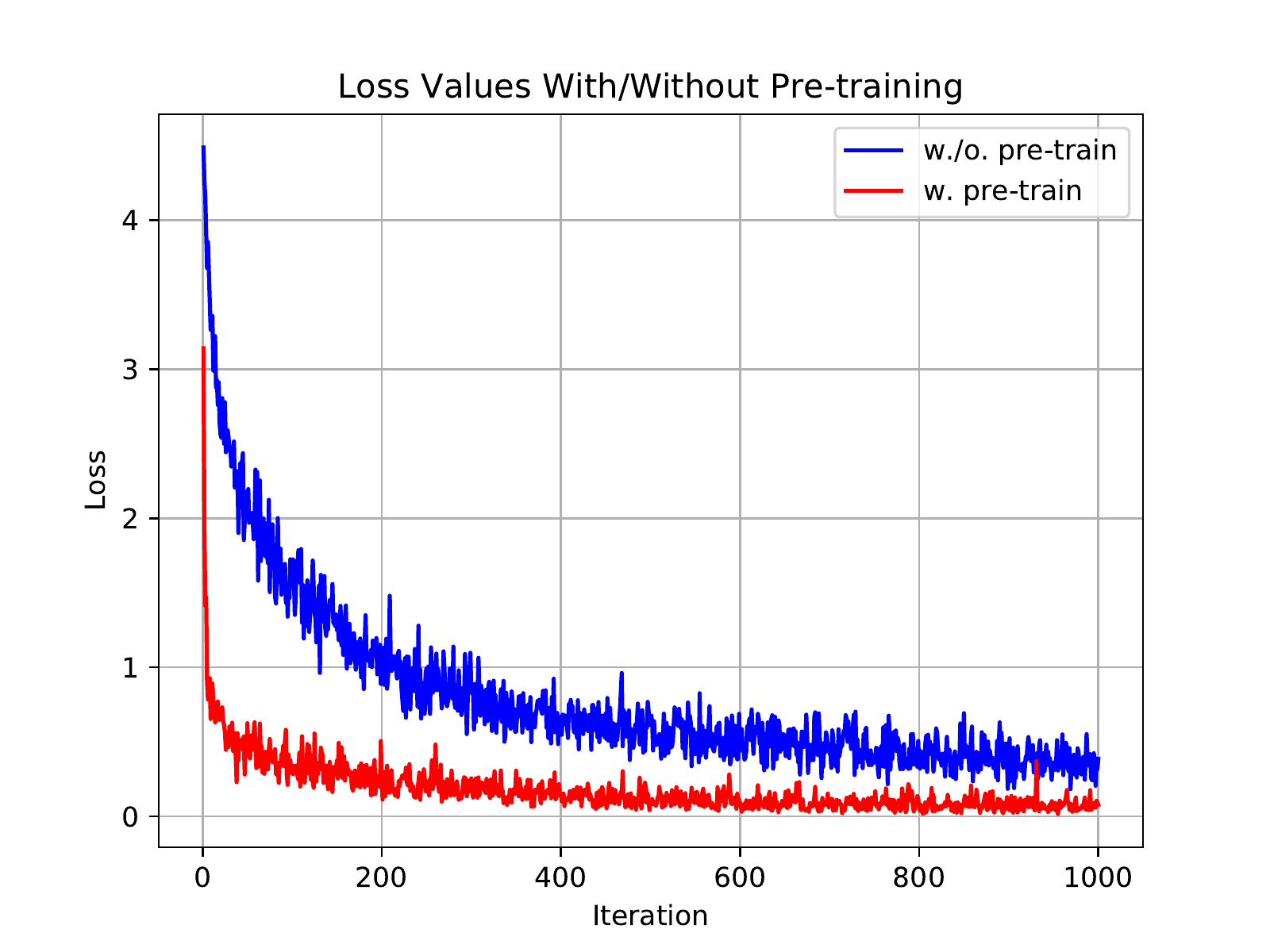} \caption{\label{figs:speed_sgm} Convergence Rate with/without Pre-training of Sketch-BERT on QuickDraw dataset.}
\end{figure}

\vspace{0.05in}
\noindent \textbf{Faster Convergence Rate of self-supervised learning by Sketch Gestalt
Model.} In addition to the improvement on classification, we also find that the pre-training sketch gestalt model can significantly
reduce the training epochs for the convergence of classification task. As the curves
shown in Fig. \ref{figs:speed_sgm}, the Sketch-BERT will converge
much faster after pre-training on Quick-Draw dataset, from about 50  to 5 epochs where one epoch has  50 Iterations in Fig. \ref{figs:speed_sgm}.

\begin{table}
\centering{}%
\begin{tabular}{c|c|c|c|c}
\hline
\multirow{2}{*}{{\small{}Models} } & \multicolumn{2}{c|}{{\small{}{}{}Classification (\%)}} & \multicolumn{2}{c}{{\small{}{}{}Retrieval(\%)}}\tabularnewline
\cline{2-5}
 & {\small{}{}{}Top-1 }  & {\small{}{}{}Top-5  }  & {\small{}{}{}Top-1 }  & {\small{}{}{}Top-5} \tabularnewline
\hline
\hline
{\small{}{}{}$345\times 70K$}  & \textbf{\small{}{}{}88.30 }{\small{}{} }  & \textbf{\small{}{}{}97.82}{\small{}{} }  & \textbf{\small{}{}{}85.47 }{\small{}{} }  & \textbf{\small{}{}{}93.49}{\small{}{} }\tabularnewline
\hline
{\small{}{}{}$345\times 5K$}  & {\small{}{}{}85.73 }  & {\small{}{}{}97.31 }  & {\small{}{}{}82.44 }  & {\small{}{}{}92.13 } \tabularnewline
\hline
{\small{}{}{}$200\times 5K$ }  & {\small{}{}{}84.89 }  & {\small{}{}{}97.14 }  & {\small{}{}{}81.87 }  & {\small{}{}{}92.07} \tabularnewline
\hline
{\small{}{}{}$100\times 5K$}  & {\small{}{}{}85.82 }  & {\small{}{}{}97.31 }  & {\small{}{}{}81.91 }  & {\small{}{}{}92.01} \tabularnewline
\hline
\end{tabular}\caption{\label{tab:res_ablation_volume} The performance of classification
and retrieval tasks of Sketch-BERT with different volumes of pre-training data.}
\end{table}

\vspace{0.05in}
\noindent \textbf{Different Volumes of Pre-training Tasks}.
We also study how the volume of pre-training data affects the downstream
tasks. We test the classification and retrieval tasks on 100 classes
with 5K training, 2K validation and 2K test samples in QuickDraw dataset.
By varying the number of classes and the number of training samples
in each class, we get different settings for pre-training tasks as
shown in Tab. \ref{tab:res_ablation_volume}. We denote the volume of pre-training data as  $c\times n$ , where $c$ is the number of classes and $n$ is the number of training samples in each class. We can find there is no obvious improvement after increasing the number of categories for pre-training data. But the number of pre-training samples in each class affects the performance in a more fundamental way, as reflected by the $3\%$ improvement on top-1 accuracy.

\vspace{0.05in}
\noindent \textbf{Sketch-BERT Architecture Analysis.}
We further compare different variants of Sketch-BERT, as shown in Tab.\ref{tab:res_ablation_struct}. We show that a reasonable depth and width of the network is important to  Sketch-BERT. Particularly,
We denote the structure of Sketch-BERT by three key hyper-parameters
$L-A-H$: number of layers $L$, number of self-attention heads $A$,
hidden size $H$. It shows that the architecture  $8-12-768$ makes a good balance between the model complexity
and final performance of Sketch-BERT model, if compared against the other variants.
When
hidden size is small,\emph{e.g.},  $H=256$, a deeper Sketch-BERT can help
increase the capacity for learning representation of sketches, clarified
by the $2\%$ improvement from $L=6$ to $L=12$ on both classification
and retrieval tasks. Nevertheless, we found the Sketch-BERT with 12 layers ($12-16-1024$) has slightly inferior results to  the other variants, and hard to get converged.

\begin{table}
\centering{}%
\begin{tabular}{c}
\hspace{-0.2in}%
\begin{tabular}{c|c|c|c|c}
\hline
\multirow{2}{*}{{\small{}Models} } & \multicolumn{2}{c|}{{\small{}{}{}{}Classification}} & \multicolumn{2}{c}{{\small{}{}{}{}Retrieval}}\tabularnewline
\cline{2-5}
 & {\small{}{}{}{}Top-1 (\%)}  & {\small{}{}{}{}Top-5 (\%)}  & {\small{}{}{}{}Top-1 (\%)}  & {\small{}{}{}{}Top-5(\%)} \tabularnewline
\hline
\hline
{\small{}{}{}{}6-8-256}  & {\small{}{}{}{}84.83}  & {\small{}{}{}{}96.42}  & {\small{}{}{}{}81.06}  & {\small{}{}{}{}91.86} \tabularnewline
\hline
{\small{}{}{}{}12-8-256}  & {\small{}{}{}{}86.34}  & {\small{}{}{}{}97.15}  & {\small{}{}{}{}83.23}  & {\small{}{}{}{}92.13} \tabularnewline
\hline
{\small{}{}{}{}12-16-1024}  & {\small{}{}{}{}85.31}  & {\small{}{}{}{}97.44}  & {\small{}{}{}{}82.76}  & {\small{}{}{}{}92.11} \tabularnewline
\hline
{\small{}{}{}{}8-12-768}  & \textbf{\small{}{}{}{}88.30 }{\small{}{}{}}  & \textbf{\small{}{}{}{}97.82}{\small{}{}{}}  & \textbf{\small{}{}{}{}85.47 }{\small{}{}{}}  & \textbf{\small{}{}{}{}93.49}{\small{}{}{} }\tabularnewline
\hline
\end{tabular}\tabularnewline
\end{tabular}
\caption{\label{tab:res_ablation_struct} The performance of classification
and retrieval tasks for different structures of Sketch-BERT ($L-A-H$).}
\end{table}

\begin{figure}
\centering{}\includegraphics[scale=0.49]{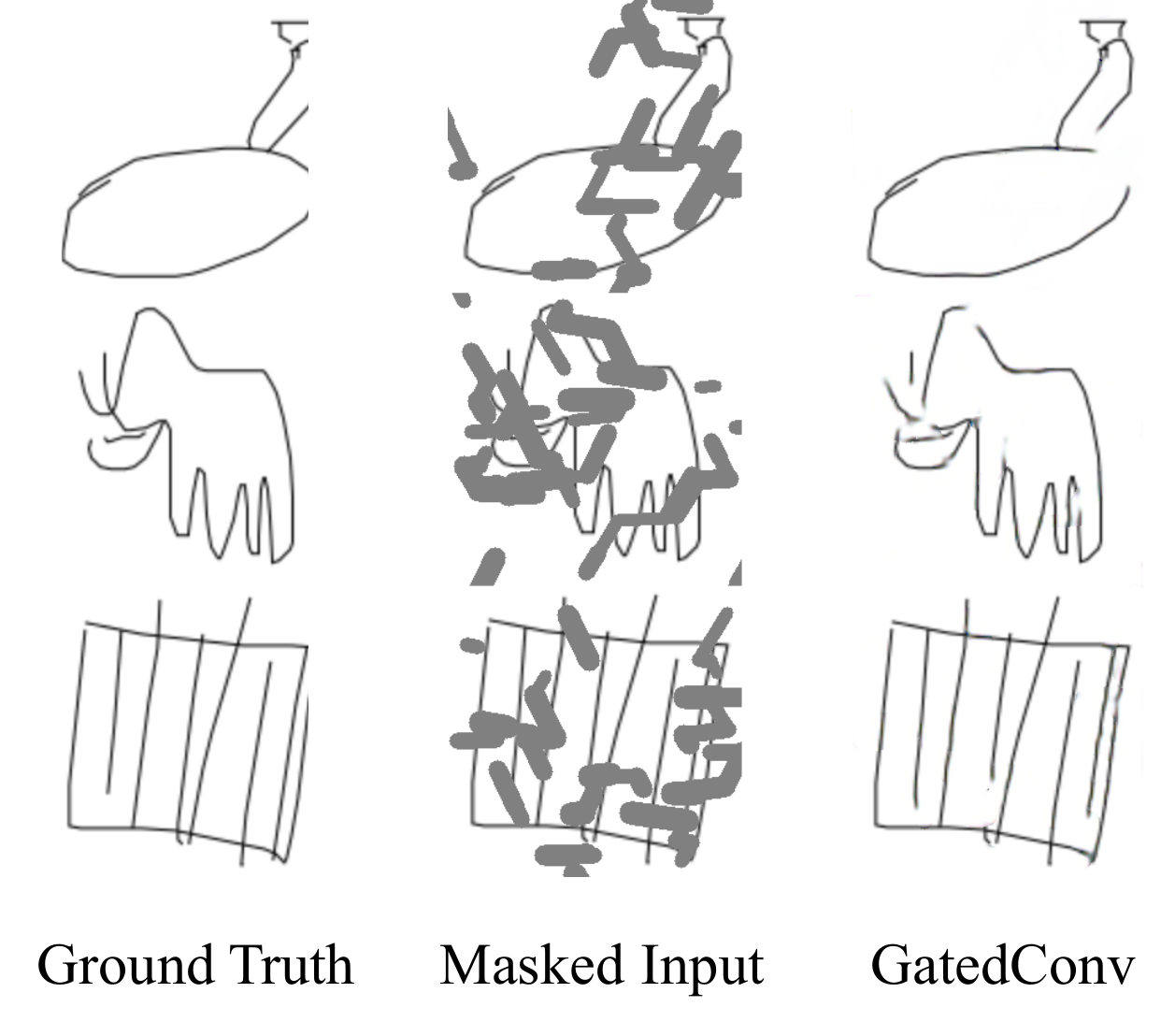} \caption{\label{figs:result_cnn_comp} CNN based models for sketch gestalt
Task. We employ Gated Convolution \cite{yu2019free} to complete the masked sketches.}
\end{figure}

\noindent \textbf{Sketch Gestalt by CNN based Model.}
We further conduct experiment to show that the proposed sketch gestalt task is very difficult.
We use the Gated Convolution \cite{yu2019free} model to train on QuickDraw dataset with random masks. It is difficult for such CNN based
model to reconstruct the shape of complicated sketches; and the results always exist artifacts. Since the different input requirement of image inpainting and sketch gestalt, the ``Masked Input" terms in Fig.~\ref{figs:result_cnn_comp} use irregular masks which is fundamentally different from the terms in Fig.~\ref{figs:result_comp}. The models
for image inpainting always aim at recovering the masked parts by borrowing
the patches from other parts of the image, while it is not tailored to
sketch gestalt. 

\section{Conclusion}

In this work, we design a novel Sketch-BERT model for sketch representation
learning which employs the efficient self-supervised learning by sketch gestalt. A novel sketch gestalt model is proposed for self-supervised learning  task of sketches. The results on QuickDraw and TU-Berlin
datasets show the superiority of Sketch-BERT on classification and retrieval tasks. We also conduct experiments
on sketch gestalt task to show the ability of Sketch-BERT
on generative representation learning. Furthermore, the Sketch-BERT model can be extended to more tasks for sketches like sketch based image retrieval and sketch generation which can be studied in future.

\section{Acknowledgements}
This work was supported in part by NSFC Projects
(U1611461,61702108), Science and Technology Commission
of Shanghai Municipality Projects (19511120700),
Shanghai Municipal Science and Technology Major
Project (2018SHZDZX01), and Shanghai Research
and Innovation Functional Program (17DZ2260900).
\bibliographystyle{ieee_fullname}
\bibliography{egbib}

\end{document}